\documentclass{article}

\usepackage{arxiv}

\usepackage[utf8]{inputenc} 
\usepackage[T1]{fontenc}    
\usepackage{hyperref}       
\usepackage{url}            
\usepackage{booktabs}       
\usepackage{amsfonts}       
\usepackage{nicefrac}       
\usepackage{microtype}      
\usepackage{lipsum}		
\usepackage{graphicx}
\usepackage{natbib}
\usepackage{doi}
\usepackage{array}

\title{Exploring Commonalities in Explanation Frameworks: A Multi-Domain Survey Analysis}

\author{ \href{https://orcid.org/0000-0002-3664-5367}{\includegraphics[scale=0.06]{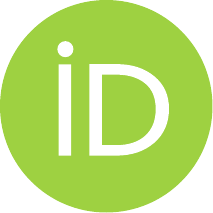}\hspace{1mm}Eduard Barbu} \\
	Institute Of Computer Science\\
	 Tartu, Estonia\\
	\texttt{eduard.barbu@ut.ee} \\
	\And
	\href{https://orcid.org/0000-0001-5414-6089}{\includegraphics[scale=0.06]{orcid.pdf}\hspace{1mm}Marharyta Domnich} \\
	Institute Of Computer Science\\
	 Tartu, Estonia\\
	\texttt{marharyta.domnich@ut.ee} \\ \\
     \AND
	\href{https://orcid.org/0000-0002-2497-0007}{\includegraphics[scale=0.06]{orcid.pdf}\hspace{1mm}Raul Vicente} \\
	Institute Of Computer Science\\
	 Tartu, Estonia\\
	\texttt{raulvicente@gmail.com} \\
    \AND
	{\hspace{1mm}Nikos Sakkas} \\
	Apintech Ltd, POLIS-21 Group\\
	 Cyprus\\
	\texttt{sakkas@apintech.com} \\
    \AND
	\href{https://orcid.org/0000-0003-4724-1322}{\includegraphics[scale=0.06]{orcid.pdf}\hspace{1mm}André Morim} \\
	LTPlabs, Avenida da Senhora da Hora,459\\
	 Porto, Portugal\\
	\texttt{andre.morim@ltplabs.com} \\
}

\renewcommand{\shorttitle}{\textit{arXiv} Template}

\hypersetup{
pdftitle={A template for the arxiv style},
pdfsubject={q-bio.NC, q-bio.QM},
pdfauthor={David S.~Hippocampus, Elias D.~Striatum},
pdfkeywords={First keyword, Second keyword, More},
}

\begin{document}
\maketitle

\begin{abstract}
This study presents insights gathered from surveys and discussions with specialists in three domains, aiming to find essential elements for a universal explanation framework that could be applied to these and other similar use cases. The insights are incorporated into a software tool that utilizes GP algorithms, known for their interpretability.
The applications analyzed include a medical scenario (involving predictive ML), a retail use case (involving prescriptive ML), and an energy use case (also involving predictive ML). We interviewed professionals from each sector, transcribing their conversations for further analysis. Additionally, experts and non-experts in these fields filled out questionnaires designed to probe various dimensions of explanatory methods.
The findings indicate a universal preference for sacrificing a degree of accuracy in favor of greater explainability. Additionally, we highlight the significance of feature importance and counterfactual explanations as critical components of such a framework. Our questionnaires are publicly available to facilitate the dissemination of knowledge in the field of XAI.
\end{abstract}


\section{Introduction}
The advent of AI solutions in various domains has prompted a significant shift towards data-driven decision-making. Despite the remarkable capabilities of ML models, their inherent complexity and lack of transparency pose challenges, particularly in sensitive sectors like healthcare, retail, and energy. This paper investigates the commonalities across different explanation types within these domains, focusing on enhancing understandability and trust among users. We identify essential features of an explanation framework that could cater to diverse use cases while promoting explainability and user trust by analyzing responses from domain experts and layman practitioners through questionnaires and interviews. Our research focuses on genetic programming (GP) within machine learning (ML) models. However, the insights we've obtained apply to any ML model category that supports interpretability.
Symbolic expressions derived from GP offer several advantages for enhancing explainability in ML, including intuitive representations easily understood by non-experts and the simplicity and compactness of solutions that capture essential patterns in the data \cite{gp_review}. Their flexibility allows adaptation to various types of problems and data, contributing to the transparency of AI systems by explicitly representing how inputs are transformed into outputs. 
The paper is organized as follows: we begin with an overview of related work. This is followed by introducing the three distinct use cases and their unique characteristics. In Section \ref{survey_methods}, we elaborate on the methodology employed in conducting the surveys. The paper concludes with a discussion of our findings and presents conclusions, including recommendations for developing a GP tool to support practitioners across three use cases.
The developed questionnaires are publicly available to facilitate the dissemination of knowledge in the field of XAI.

\section {Related Work}

There is a solid body of research encompassing various perspectives and methodologies related to user expectations from AI system explanations. For example, \cite{LANGER2021103473} outlines the key stakeholder groups advocating for the explainability of the AI systems and evaluates their requirements. It presents a framework detailing essential concepts and relationships to assess, tailor, select, and develop solutions to meet stakeholder needs.

Researchers have focused on understanding the human aspect of XAI by developing reliable tools to collect opinions from experts, and the public \cite{Gunning_Aha_2019,gunning2021darpa}.

In assessing explanations in ML system user surveys, the System Causability Scale is a notable tool that consists of a ten-question survey using a 5-point Likert scale (ranging from strongly disagree to strongly agree). It is used to evaluate the usability of user interfaces that present ML model explanations, and it was introduced in \cite{DBLP:journals/corr/abs-1912-09024}.  Similarly, the System Usability Scale uses a ten-item Likert questionnaire to gauge user feedback on the effectiveness of generated explanations \cite{DRAGONI2020101840}. On the other hand, \cite{hoffman2023measures} suggests the Explanation Goodness Checklist, aimed at researchers for evaluating explanation properties, and the Explanation Satisfaction Scale, designed to capture users' appraisal of explanations. These measures, grounded in cognitive psychology and the philosophy of science, assess essential qualities that constitute a good explanation.

The insights from psychometrics have been leveraged to create and assess a new human-centered questionnaire designed to evaluate explanations generated by XAI methods reliably \cite{10.1007/978-3-031-44070-0_11}. Recognizing that explainability encompasses several dimensions, this approach aims to comprehensively address the complexity of the concept within the context of AI systems.

In their literature review, the authors in \cite{agenda} define five primary goals for AI system interactions with end users: understandability, trustworthiness, transparency, controllability, and fairness. They recommend designing XAI systems to achieve these objectives and suggest guidelines for creating explanations focusing on crucial system components. Additionally, they highlight the necessity for compromises in AI explanations, underlining the absence of a one-size-fits-all solution.

Our main contribution is highlighting key elements of explanations across various fields, such as medicine, retail, and energy, aiming to create an AI tool versatile enough for professionals in diverse domains. This cross-domain approach seeks to provide a comprehensive solution applicable in multiple fields, bridging the gap between different practitioners' needs.
\section{The use cases} \label{use_cases}

This section introduces the three use cases, emphasizing their distinctive features. Two are geared toward prediction, while the third adopts a prescriptive approach.

\subsection{Medical scenario}

The medical scenario investigates explanations for paraganglioma and diabetes. The focus on paraganglioma, a rare tumor of the autonomic nervous system, highlights the challenge of its unpredictable progression. Treatment options vary widely, from watchful waiting to surgery or radiotherapy, depending on the tumor's growth. We aim to develop a GP model to assist physicians in deciding when to treat by offering predictions on tumor development and potential complications. This model intends to facilitate shared decision-making between clinicians and patients, optimize treatment schedules, minimize unnecessary interventions, and tailor patient monitoring without replacing clinical judgment.

This approach does not replace clinical decision-making but enriches it with valuable data-driven insights, aiding clinicians in navigating the complexities of paraganglioma treatment planning.

The diabetes scenario utilizes the well-known dataset \cite{smith1988using}, employed by various machine learning models, to determine the presence or absence of diabetes in a patient to build a questionnaire for diabetes prediction. This is necessary at this stage because there are no prediction models for paraganglioma.

\subsection{Retail use case} 

Grocery retailers aim to balance operational efficiency with customer satisfaction in their home delivery services, offering flexible delivery windows while keeping costs low. Dynamic Timeslot Pricing addresses the challenge of aligning customer preferences with logistical capabilities, utilizing AI to ensure transparent and fair delivery pricing and scheduling. This approach involves:

\begin{enumerate}
    \item Utilizing customer and logistics data to train models that predict customer willingness to pay and the cost-to-serve for different time slots.
    \item Developing an algorithm to harmonize customer demands with logistical efficiency, guiding timeslot and pricing strategies.
\end{enumerate}

This strategy, supported by a prescriptive model combining customer willingness to pay and cost-to-serve insights, seeks to optimize profit margins and enhance service fairness. Regular updates and consultations ensure the model's relevance and effectiveness in a real-world setting.

The heuristic, which determines slot-price configurations using a symbolic formula (Prescriptive Model), relies on two auxiliary models—the Willingness to Pay (WTP) and Cost to Serve (CTS) models—for its predictive insights. 

\subsection{Energy use case}

This use case aims to forecast home energy use, focusing on critical elements like weather conditions, past energy patterns, building dynamics, pricing schemes, and indoor temperatures to suggest savings. The challenge is to provide clear explanations for users, facilitate decision-making, and integrate these forecasts into business practices to enhance energy efficiency and decision support. The following key factors are considered:

\begin{enumerate}
    \item Weather conditions like hourly outdoor temperature and cloud coverage are vital for modeling energy usage.
    \item Historical consumption data, providing insights into usage patterns and the impact of weather changes on energy demand.
    \item Building dynamic profiles, reflecting the electrical infrastructure and consumption behaviors, informing system adjustments.
    \item The current pricing scheme is crucial for implementing demand response strategies to influence consumption based on pricing.
    \item Indoor temperature monitoring to suggest energy-saving measures.
\end{enumerate}

The core challenge lies in providing users with local explanations to facilitate informed decision-making and seamlessly integrating this solution into real-world operations. This approach emphasizes the practical application of forecasting in enhancing energy efficiency and operational decision-making.

\section {Survey methods} \label{survey_methods}

This section outlines the survey methodologies applied to the three investigated use cases. Our approach incorporated two methods: conducting interviews with domain experts and distributing questionnaires to practitioners who may not have expert knowledge.

Details of the surveyed experts are available at this link: \href{https://drive.google.com/file/d/1H9MGdjEitCMszqGJQF5Ozk8vz9-tMgbB/view?usp=sharing}{Interviewed Experts Document}. Links to the questionnaires can be found in the following subsections.

\subsection {Survey methods for the Medical Scenario}\

The development of the questionnaire for the medical use case was based on specific criteria aimed at understanding doctors' needs and preferences regarding model explanations. The target audience comprised doctors, who required a general overview and detailed technical insights into how AI model features influence predictions. Because the problem is much better studied and understood, the questionnaire focused on diabetes risk estimation. This strategy leveraged existing patient data on high diabetes risk, thereby avoiding any biases in doctors' explanation preferences while still gathering relevant insights.

The questionnaire explored various aspects of model explanations, focusing on.

\begin{enumerate}
    \item \textbf{Accuracy vs. Explainability:} This section explores doctors' willingness to prioritize model explainability over performance.
    \item \textbf{Presentation Formats:} Various formats for illustrating model logic were evaluated for their understandability and effectiveness, including:
    \begin{itemize}
        \item Symbolic Regression Graphs: Comparison of two graphs (Graph A and Graph B) illustrating model behavior, with Graph B being more complex.
        \item \textit{Protocols from Genetic Programming:} Described two protocols, one simpler with three rules and one cause per rule, and another more complex with four rules and multiple causes per rule.
        \item SHAP \cite{lundberg2017unified} Feature Importance Graphs: Displayed the contribution of patient characteristics (like gender, age, obesity) to the model's decision, highlighting feature importance and data distribution.
        \item Coefficients Table: Provided insights into feature importance through numerical coefficients in a regression model, indicating how each characteristic influences diabetes risk predictions, including the significance of coefficients and p-values.
        \item Textual Explanations: Evaluated for offering concise explanations in various forms, including causal, counterfactual, and contrastive explanations.
    \end{itemize}
\end{enumerate}

According to the document linked previously, two medical doctors have completed the questionnaires.
Participants were asked to rate each explanation format on a scale from 1 to 5 for both interpretability and effectiveness, where one indicated the lowest and five was the highest score. Interpretability ratings ranged from not interpretable (1) to very intuitive (5), while effectiveness ratings assessed how well the explanations aided in decision-making, from not helpful (1) to very effective (5).

The questionnaire for the medical scenario can be explored here:  \href{https://drive.google.com/file/d/1AZYatgHSJPI4H746H_njiel5IlqW6C9X/view?usp=sharing}{Diabetes Questionnaire} 

The interview was recorded and transcribed, focusing on the paraganglioma case.  Several key issues were explored, including identifying tumor-indicative clues in medical images, the application of statistical models for predicting tumor growth, the role of genetic factors in tumor evolution, the training protocol specialists use for new doctors in paraganglioma cases, the expectations doctors have from an AI tool in such cases, and the critical need and specifics of explanations required for understanding paraganglioma.

\subsection {Survey methods for the Retail Use Case}

The following types of questions were asked in the questionnaire for the retail use case:
\begin{enumerate}
    \item Price breakthrough. The goal of price breakthrough is to ascertain whether key aspects of the retail case, such as location, demand, basket contents, and the relevance of explanations, hold significance for the customer.
    \item Explanation Preferences: These questions evaluate the customer's preferred type of explanation and capability to comprehend the provided explanation.
    \item Summarization Assessment. This evaluates the necessity for a concise summary explanation in addition to individual item price breakdowns while examining the clarity and impact of the overall explanation on the customer.
\end{enumerate}

Like in the medical scenario, participants were prompted to evaluate each explanation format on a scale of 1 to 5, covering both interpretability and effectiveness. Here, a score 1 represents the minimum, and 5 denotes the maximum achievable score. Interpretability scores varied from not interpretable (1) to highly interpretable (5). On the other hand, effectiveness scores gauged the extent to which the explanations facilitated decision-making, from not at all helpful (1) to extremely effective (5). 
For this use case, two questionnaires have been devised for two categories of users. 

\begin{enumerate}
\item \textbf{Decision-makers} Seek a comprehensive understanding of feature contributions to model predictions for system optimization. With their expert background, they prefer detailed, technical explanations to build trust and validate the model's use based on its accuracy. \href{https://drive.google.com/file/d/1nZvkWBfNEWDAL6RsW2he0nGLrakpJpFO/view?usp=sharing}{Decision-Makers Questionnaire} 
\item \textbf{Customers} Favor straightforward, accessible explanations that still convey essential information, aiding in understanding the rationale behind received offers without overwhelming technical detail.\href{https://drive.google.com/file/d/1YeDNk3kcyhB_r9zSBE2hJIoZP3VWyCJL/view?usp=sharing}{Customers Questionnaire} 
\end{enumerate}

The interview, which was recorded as a video file, delved into issues such as finding a balance between accuracy and explainability in e-commerce models, the incorporation of graphs and mathematical formulas into explanations, understanding customer behavior through the dynamic relationship between slot availability and pricing, and designing a dynamic dashboard to manage the interaction between operational efficiency and customer behavior effectively.

\subsection {Survey methods for the Energy Use Case}
The questionnaire aims to select preferred explanation formats like tables, charts, interactive graphics, and text, as well as explanation types such as causal, contrastive, and counterfactual. We've made some assumptions to streamline our study and keep the questionnaire manageable. We focus on operational managers as the primary audience for explanations, considering their ability to provide insightful feedback. These managers seek an in-depth understanding of how different features impact model predictions and identify optimization strategies. Their expertise allows them to comprehend more technical and sophisticated explanations. The objective is to persuade decision-makers to endorse and apply the model, assuming it's sufficiently accurate, by delivering explanations that enhance their trust in the predictions. Conversely, customers are expected to favor straightforward, non-technical explanations that help them grasp and accept the reasoning behind the offers they get.

The energy questionnaire addressed questions in several key areas:
\begin{enumerate}
    \item \textbf{Accuracy-Explainability Trade-off}: Explores the willingness of operational managers to sacrifice some model accuracy in favor of increased explainability.
    \item \textbf{Importance of User Explanations}: Determines how much value users place on explanations provided by the system regarding forecasting operations, their necessity in various customer scenarios, and preferences for the types and frequency of explanations.
    \item \textbf{What-if Explanations (Counterfactuals)}: Assesses the importance of explaining to users how their actions could impact forecasting outcomes, the significance of these actions in different scenarios, the preferred forms of what-if explanations, and how often they should be provided.
    \item \textbf{Facility Managers' Explanations}: Evaluates the need for comprehensive explanations for facility managers regarding data and models for all unique building spaces, including the relevance of what-if explanations and the importance of visualizing feature importance through methods like SHAP feature importance graphs.
\end{enumerate}

All interviewed experts and five additional energy experts have completed the questionnaire. 
\href{https://drive.google.com/file/d/1RtqPGbYGW2wgQgoAqzbNSkgLfSpeoDc2/view?usp=sharing}{Energy Questionnaire} 

The interviews delved into the energy solution from various angles, each tailored to the interviewee's expertise. Discussions ranged from addressing market challenges in energy solutions and the importance of clear explanations for end-users to exploring energy consumption disaggregation and the role of genetic programming in enhancing analysis. Insights were also shared on leveraging machine learning for water consumption monitoring to optimize resource management and identify inefficiencies. Additionally, the design and usability of user interfaces for energy management systems were examined, emphasizing the need for intuitive and engaging interfaces to manage energy consumption better.

\section {Results} \label{results}

\subsection{Medical scenario}
\begin{figure}[ht]
\centering
\includegraphics[width=0.8\textwidth]{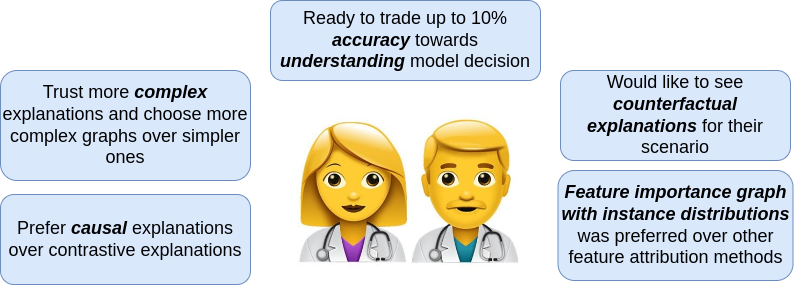}
\caption{Insights into doctors' preferences for medical case derived from the questionnaire.}
\label{fig:insights-diabetes}
\end{figure}

Figure \ref{fig:insights-diabetes} summarizes key findings from the diabetes questionnaire.

The doctors' preferences for AI system explanations' key findings include a willingness to trade off a small percentage of model accuracy for greater explainability, challenges in understanding complex graphical representations like graphs and symbolic regression, and a preference for detailed yet intuitive explanations such as protocols and SHAP feature importance graphs. Simplification and clarity were highlighted as essential for effectively conveying model logic, with counterfactual explanations being particularly valued for their potential to improve patient understanding and therapy compliance.

 Feature importance graphs were most favored, followed by textual explanations and rule-based protocols. Graphs and coefficient tables were least preferred due to concerns about understandability.

 Key insights from the interviews have illuminated several important aspects. The efforts in developing models for paraganglioma cases are pioneering, as there are no existing benchmarks to measure the accuracy of our models. The potential influence of genetic information on the development and progression of paraganglioma was recognized, underscoring the importance of personalized medicine. Additionally, a notable deficiency in current medical practices is the absence of tools for effectively monitoring tumor growth, indicating a crucial area for technological development. Furthermore, doctors are seen to value the predictions provided by our models, integrating them into patient communications. This underscores the need for models to offer numerical predictions and provide explainable insights that enhance doctor-patient trust and decision-making processes. 
 Initial GP models for paraganglioma have undergone testing. For those interested in examining the outcomes, references \cite{sijben2024function} offer detailed insights into the results.

\subsection{Retail use case}

\begin{figure}[ht]
\centering
\includegraphics[width=0.8\textwidth]{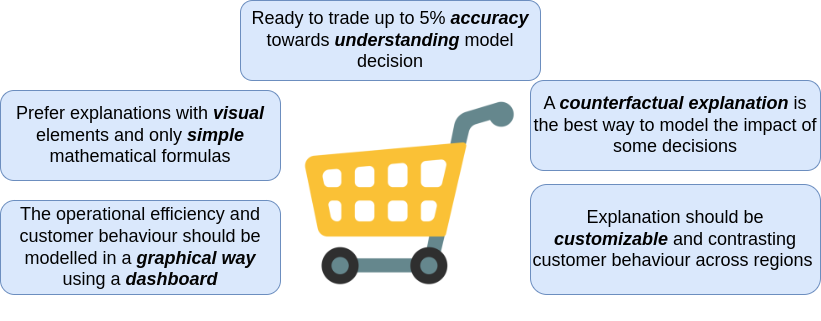}
\caption{Insights into online retail decision-makers preferences derived from the questionnaire.}
\label{fig:insights-retail}
\end{figure}

The decision-makers seek explanations across various dimensions: customer behavior, transportation costs, and strategies for maximizing profits. The questionnaires findings are summarized in the figure \ref{fig:insights-retail}

In feedback from decision-makers on AI system explanations, there's an openness to sacrificing a portion of model performance for enhanced explainability, with preferences for detailed yet intuitive insights into model workings. This encompasses a broad interest in customer behavior, cost analysis, and profit strategies, highlighting a desire for interactive tools and visualizations that facilitate deeper understanding and strategic adjustments. There's a notable emphasis on practical application, with decision-makers valuing features like counterfactual explanations and the ability to interpret and act upon complex information, all aimed at optimizing operational efficiency and customer engagement.
The interview yielded several important insights. There's an openness to sacrificing accuracy for enhanced explainability, although her limited expertise in machine learning cautions against hasty decisions on F1 score trade-offs. The preference leans towards using visual elements in explanations, carefully including simple mathematical formulas to avoid confusion with complex operations like logarithms. Using graphical dashboards is recommended for analyzing operational efficiency and customer behavior, as they improve both interpretability and user interaction. Counterfactual explanations are particularly valued for their ability to illustrate the outcomes of specific decisions, like the introduction of new scheduling slots. Developing models delineating customer characteristics by region and differentiating behaviors across these regions are highlighted as essential for gaining deeper business insights.

\subsection{Energy use case}

\begin{figure}[ht]
\centering
\includegraphics[width=0.8\textwidth]{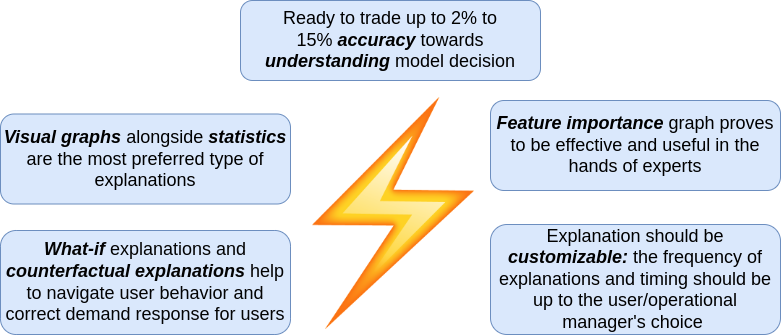}
\caption{The insights from the energy questionnaire from operational and facility managers}
\label{fig:insights-energy}
\end{figure}

The insights from operational and facility managers are summarized in figure \ref{fig:insights-energy}.

Operational managers are receptive to balancing accuracy with increased transparency, showing flexibility in the accuracy-explainability trade-off, with the degree of compromise varying based on the audience. Preferences lean towards visual and simple mathematical explanations to cater to different technical levels among stakeholders. Insights into operational efficiency and customer behavior are effectively conveyed through graphical dashboards, while counterfactual explanations offer valuable scenario analysis tools. The approach includes strategic analyses like regional customer behavior modeling and what-if scenarios, underscoring the utility of feature importance graphs and counterfactual explanations for providing clear and actionable insights across various aspects of decision-making and management.
Insights from the interviews suggest that a forecasting solution incorporating explanations is favored over a basic model. The methodologies used in forecasting and providing explanations are also applicable to other sectors, including gas and energy consumption. It's recommended that the graphical interface for end-users be designed for ease of use, possibly including interactive features such as knobs. Additionally, for energy-related applications, incorporating a smartphone component in the tool solution to send convenient notifications to users is advised.
For those interested in exploring the performance of GP models within the energy use case, further information and detailed analyses are available in references \cite{nikos1} and \cite{nikos2}.

\subsection {General guidelines}

The table \ref{tab:guidelines_insights} summarizes the overarching guidelines derived from the survey findings.

\begin{table}[ht]
\centering
\caption{Guidelines and Insights from User Studies on Explanatory Tool's Architecture}
\label{tab:guidelines_insights}
\begin{tabular}{>{\raggedright\arraybackslash}p{0.15\linewidth} >{\raggedright\arraybackslash}p{0.4\linewidth} >{\raggedright\arraybackslash}p{0.35\linewidth}}
\toprule
\textbf{Domain} & \textbf{Insight} & \textbf{Recommendation} \\
\midrule
All & Preference for explainability over perfect accuracy, feature importance graphs as effective communication tools, and value of counterfactual explanations. & Balance explainability and accuracy, utilize feature importance graphs, and supplement counterfactuals for comprehensive understanding. \\
\bottomrule
\end{tabular}
\end{table}

Drawing from these insights, the design of the explanatory tool should incorporate two essential modules: a Counterfactual Module, which calculates the minimal changes required to shift the model's decision towards a desired outcome, thereby enabling "What-if" scenarios based on user queries, and a Global Importance Module, which provides visualization of the significant feature contributions to the model's predictions, in line with findings from the user studies. Both modules should be integrated within the tool, ensuring that the inputs, outputs, and connections between modules are well-defined.

\section{Conclusions}
This study identifies foundational components for an XAI framework intended for various applications through comprehensive questionnaires and interviews with domain experts in three distinct use cases. The envisioned XAI tool incorporates a Counterfactual Module to facilitate "What-if" scenarios, allowing users to see how minimal changes could lead to desired outcomes. Additionally, a Global Importance Module is designed to visually represent the most influential features in model predictions, resonating with the XAI literature emphasizing the critical role of feature importance and counterfactual explanations. While aiming for shared applicability, the framework also acknowledges the unique requirements of each specific case, although the detailed exploration of these unique case aspects was beyond this paper's scope. 
This approach informs the ongoing development of the AI tool, leveraging insights gathered from user studies to ensure the tool's effectiveness across different domains.
For future research, the interest in online retail and energy sectors for customizable and user-specific explanations points towards a growing trend. This trend leans towards integrating NLP interactivity into explanations, an area we are beginning to explore.

\section{Acknowledgments}
  This research was conducted under the Transparent, Reliable, and Unbiased Smart Tool for AI (Trust-AI) project, with Grant Agreement ID: 952060, funded by the EU Commission.

\bibliographystyle{unsrtnat}
\bibliography{commonalities}  

\end{document}